\documentclass[letterpaper, 10 pt, conference]{ieeeconf}  
\IEEEoverridecommandlockouts                              
\usepackage{graphicx} 
\usepackage{comment}
\graphicspath{{./figures/}}

\usepackage{amsmath} 
\usepackage{amssymb}  
\usepackage{cite}
\usepackage{units}
\usepackage{bm} 
\usepackage{multirow}
\usepackage{todonotes}
\usepackage[hidelinks]{hyperref}
\usepackage[capitalise]{cleveref}

\usepackage{tabularx}

\usepackage{algorithm}
\usepackage{algpseudocode}

\makeatletter
\let\NAT@parse\undefined
\makeatother

\crefname{equation}{}{}
\crefname{figure}{Fig.}{Figs.}

\graphicspath{{./Figures/}}
\usepackage{xcolor}
\usepackage{makecell}
\usepackage{gensymb}

\title{\LARGE \bf
Dynamic Bipedal Maneuvers through \\ Sim-to-Real Reinforcement Learning

}

\author{Fangzhou Yu, Ryan Batke, Jeremy Dao, Jonathan Hurst, Kevin Green, and Alan Fern 
\thanks{*This work is supported by the NSF Grants No. IIS-1849343, NSF Grants 1314109-DGE and DARPA Contract W911NF-16-1-0002.}
\thanks{All authors are with Collaborative Robotics and Intelligent Systems Institute, Oregon State University, Corvallis, Oregon, 97331, USA. Email: \{{\tt\footnotesize yufangzh, batker, daoje, jonathan.hurst, greenkev, alan.fern}\}@oregonstate.edu. }
}

\begin{document}

\maketitle
\thispagestyle{empty}
\pagestyle{empty}

\begin{abstract}

For legged robots to match the athletic capabilities of humans and animals, they must not only produce robust periodic walking and running, but also seamlessly switch between nominal locomotion gaits and more specialized transient maneuvers. 
Despite recent advancements in controls of bipedal robots, there has been little focus on producing highly dynamic behaviors.  
Recent work utilizing reinforcement learning to produce policies for control of legged robots have demonstrated success in producing robust walking behaviors. 
However, these learned policies have difficulty expressing a multitude of different behaviors on a single network.
Inspired by conventional optimization-based control techniques for legged robots, this work applies a recurrent policy to execute four-step, 90\degree{} turns trained using reference data generated from optimized single rigid body model trajectories.
We present a novel training framework using epilogue terminal rewards for learning specific behaviors from pre-computed trajectory data and demonstrate a successful transfer to hardware on the bipedal robot Cassie.
   
\end{abstract}

\section{INTRODUCTION}

Animals exhibit a multitude of dynamic behaviors, such as squirrels leaping from treetops and birds taking off and landing. Moreover, they are also able to seamlessly transition between behaviors.
Robots that match human and animal athletic capability will require a control architecture that enables them to transition between behaviors with the same fluidity. 
As legged robots evolved over the past few decades to become more agile and dynamic, their control algorithms became more sophisticated to take advantage of advances in mobile computing \cite{DesignDynLegRob}. 
Recent developments in the realm of legged locomotion controls prominently feature the use of model predictive control (MPC) and optimization techniques to generate trajectories for quadrupedal jumps and backflips \cite{cheetahAutoJump,backflip_mini}. 
Using RL to train neural network locomotion controllers has also shown to be a promising alternative avenue of research, enabling Cassie, another human-scale bipedal robot, to perform dynamic gaits ranging from walking, running, skipping, and stair climbing on real-world hardware \cite{all_gaits, blind_stairs}. 
This work extends upon the periodic reward composition method of learning bipedal gait policies for Cassie \cite{all_gaits} by including trajectory data derived offline using a single rigid body model (SRBM) \cite{Batke2022SRBM} with the aim of producing policies that learn to transfer the trajectories of the reduced-order model (ROM) to Cassie with its full-order dynamics in the real world. 
The challenge of successfully transitioning between different policies is addressed during the training process with the concept of an epilogue terminal reward. To prove the viability of our proposed technique implemented on real world hardware, we demonstrate the successful sim-to-real transfer of Cassie performing a four-step 90 degree right turn using a policy trained with trajectory data that successfully transitions between another policy performing a running gait developed from our previous work. 
\begin{figure}
  \centering
      \includegraphics[scale=0.2]{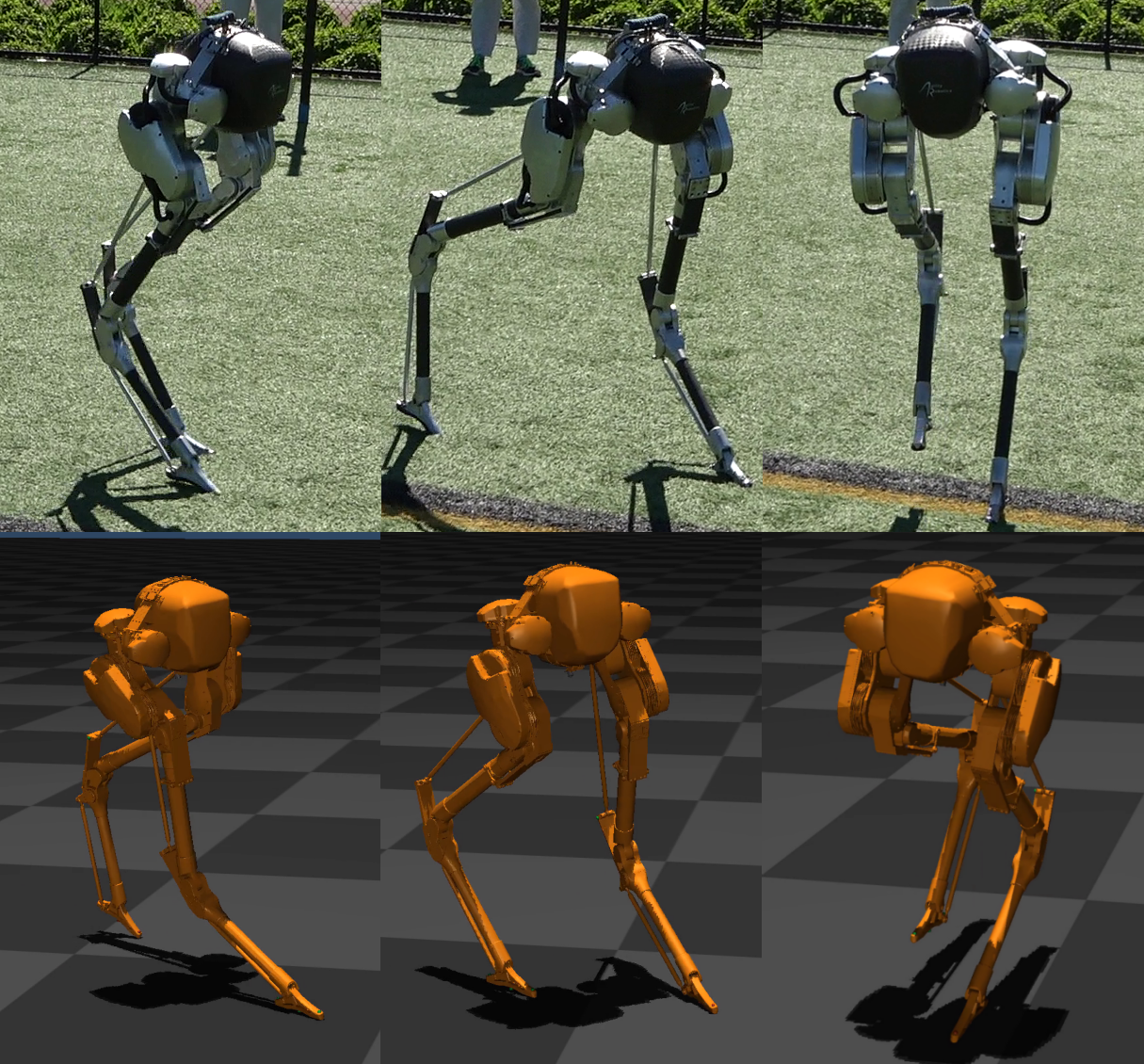}
      \vspace{-14 pt}
  \caption{A Cassie robot executing a four-step 90\degree{}  right turn. \textbf{(Top Row)} Hardware field test of the full-reference turning policy initialized from a commanded heading speed of 2.0m/s on artificial turf. \textbf{(Bottom Row)} Cassie running the full-reference turning policy in simulation initialized from a target heading speed of 2.5 m/s.}
  \label{fig:lead}
\end{figure}
\section{Related Works}
\subsection{Learning for General Locomotion} \label{learn-gen-locomotion}

RL has shown to be a promising alternative to model-based control of legged robots \cite{all_gaits,AgileSim2Real,anymalRL}. However, most published work on the application of RL to legged locomotion focuses on performing cyclic gaits, while in this work we are concerned with more dynamics one-off maneuvers. Prior work on performing different behaviors with learned methods use RL to train a singular policy to execute all desired behaviors instead of training separate policies for each individual behavior. This causes the behavior space of the policies to be limited by the richness of a singular reward function, making them ill-suited to learning multiple different behaviors. This work addresses this issue by switching between policies trained to execute specific behaviors.
\cite{all_gaits} demonstrated a learning framework capable of reproducing all common bipedal gaits for Cassie on a single policy that does not use expert reference trajectories. The resulting policy could continuously transition between different bipedal gaits by adjusting a left and right foot cycle offset parameter. More recent work \cite{helei_stepping_stones} used a similar learning framework to the one in this work to develop policies for Cassie to step on target footholds. The foothold targeting policies were allowed to modulate the gait behavior to achieve strides of varying length by adaptively changing the stepping frequency. Model differences between simulation and the real world robot were resolved by training the policy with dynamics randomization \cite{AgileSim2Real}, which has shown to improve the consistency of successful sim-to-real transfers of recurrent neural network (RNN) walking policies for Cassie \cite{LSTMCassie}. 

\subsection{Learning for Multiple Behaviors}

RL has also been used to learn different locomotion skills as well as smooth transitions between them. 
\cite{DeepMindLearnCoreSkills} demonstrated successful sim-to-real transfer of switching between forward and backward walking on a single policy trained with atomic, task-specific reward functions. Similarly, learned locomotion control policies have shown to be capable of assuming different gait behaviors to negotiate terrain obstacles and gaps. \cite{deepmind_learn_locomotion_rich_envs} demonstrated the emergence of robust obstacle clearing behavior for torque-controlled legged agents by training control policies using simple reward functions in obstacle-rich simulation environments. These examples of prior work engineer the agent-environment interactions to  encourage the emergence of multiple behaviors, which suffer from the same drawbacks as the examples in \cref{learn-gen-locomotion} because multiple behaviors are learned on a single policy.

\subsection{Model Based Methods}

\emph{Trajectory optimization (TO)} has been a hallmark feature in motion planning for modern dynamic legged locomotion \cite{cfrost,backflip_mini,posa,Kuindersma2016OptimizationbasedLP}. 
In this context, trajectory optimization is a tool used to yield a plan for future robot states given an initial state, such as contact wrenches and \emph{center-of-mass (CoM)} positions. 
The fidelity of the model used for TO varies from detailed, full-order dynamic and kinematic representations of actual hardware \cite{posa} to reduced-order dynamic, full-order kinematic models \cite{DaiWBC_centroidal}, down to minimal centroidal/SLIP models amenable for MPC \cite{Apgar2018FastOT,cheetahAutoJump}. 
In this work, we choose to use the single rigid body model for its ability to capture linear and rotational dynamics while being easy to describe mathematically. 
TO-based control techniques for legged robots have also shown success in composing behaviors in recent work. \cite{cheetahAutoJump} used a MPC strategy to execute running leaps for the Cheetah 2 robot over obstacles.
\cite{QuanJump} used offline TO to generate jumping trajectories and a separate MPC style landing controller that targeted an optimal distribution of foot contact forces to perform jumps and successful landings for Cheetah 3. 
\cite{hutterComp} achieved smooth transitions between a large variety of different behaviors by using TO to generate a pre-computed motion library, and targeted sequences of desired library motions using MPC that plan over shorter time horizons. This approach is similar to the approach taken by Boston Dynamics \cite{BD_parkour} for recent work on the Atlas bipedal robot. Model-based methods are capable of producing complex, dynamic behavior on legged robots, but they are challenging to implement. To the best of our knowledge, model-based methods have yet to demonstrate highly dynamic behaviors on bipedal robots within the academic research community. In comparison, training control policies using RL algorithms are a more straightforward method of transferring dynamic motion plans to controllers for the full-order robot. 

\subsection{Learning with Trajectory Optimization}

Methods attempting to combine learned policies with techniques stemming from model-based legged locomotion controls are also seen in existing literature. \cite{DeepGait} used a linear program (LP) with a centroidal dynamics robot model from \cite{CCROC} to assess the feasibility of transitioning between foothold candidates. This feasibility assessment was used as a termination condition in the training of a gait planning policy for the quadrupedal ANYmal robot. Use of TO data have also been used to directly shape the rewards for RL locomotion control policies. \cite{deepmind_centroidal} leveraged TO to generate target reference trajectories over difficult terrain using a centroidal dynamics model, and used the data from TO to train RL control policies using imitation learning to transfer the terrain traversing gait plan to the full-order ANYmal robot model in simulation. Prior works in this area have not yet applied the technique of learning control policies from TO data to perform highly dynamic behaviors such as high speed turning, which is what we investigate in this work.  

\section{Methods}

A four step 90\degree{} right turn was selected as the target behavior for the control policies in this work because its aperiodic and highly dynamic nature marks a significant departure from the dynamical regime of regular walking. 
Control policies trained to execute the turning behavior are initialized from states derived from a pretrained walking policy and transition back to the walking policy at the end of the turning maneuver.  
Matching the terminal state of the reference trajectory is not guaranteed to permit successful transitions back to walking policies, so turning policies must learn how to deviate away from tracking the reference data to facilitate successful transitions. 
This challenge is addressed using epilogue rewards detailed in \cref{epilogue-method}, and is a novel component of our proposed learning framework.  
We train recurrent control policies to perform four-step, 90\degree{} turns using Proximal Policy Optimization (PPO) in simulation \cite{ppo}.
The simulator we use is MuJoCo, extended with the robot's state estimator and noise models\footnote{Simulation available at \href{https://github.com/osudrl/cassie-mujoco-sim}{https://github.com/osudrl/cassie-mujoco-sim}}.
Previous work has shown that highly accurate simulations such as this are effective at producing control policies that transfer to hardware with no additional adaptation \cite{jeremyLoad,all_gaits,DeepMindLearnCoreSkills}.

\subsection{Reference Trajectory Optimization}
Dynamic legged maneuvers require abrupt changes in linear and angular momentum while heavily constrained by underactuation constraints.
We hypothesize that in these contexts reference information could be more useful than it was previously shown to be in nominal, steady-state locomotion.
To provide a rich library of reference motions we perform trajectory optimization with an SRBM, representing a reduced-order model of locomotion.
The SRBM approximates the complex multibody dynamics of a robot into a single rigid-body with dynamics that are manipulated via ground reaction forces applied at footholds.
We apply a widely-used, prescribed contact sequence, direct collocation trajectory optimization method \cite{coalesce}.
This allows the optimization to adjust foot timings, but not the sequence of contacts.
This is not overly restrictive as bipedal robots have only a small space of feasible contact patterns.
Our contact pattern for four-step turns is a grounded run consisting of alternating phases of single-stance with instant transfer.
We apply a set of transferability constraints which ensure the resulting trajectories are more directly applicable to the target Cassie robot.
These include maximum ground reaction force, friction cones, maximum yank (time rate of change of force), leg length limits, and foot placement constraints to prevent leg crossing.
More details on the library generation method can be found in \cite{Batke2022SRBM}.

\begin{figure}
  \centering
  \vspace{4pt}
  \includegraphics[width=.9\columnwidth]{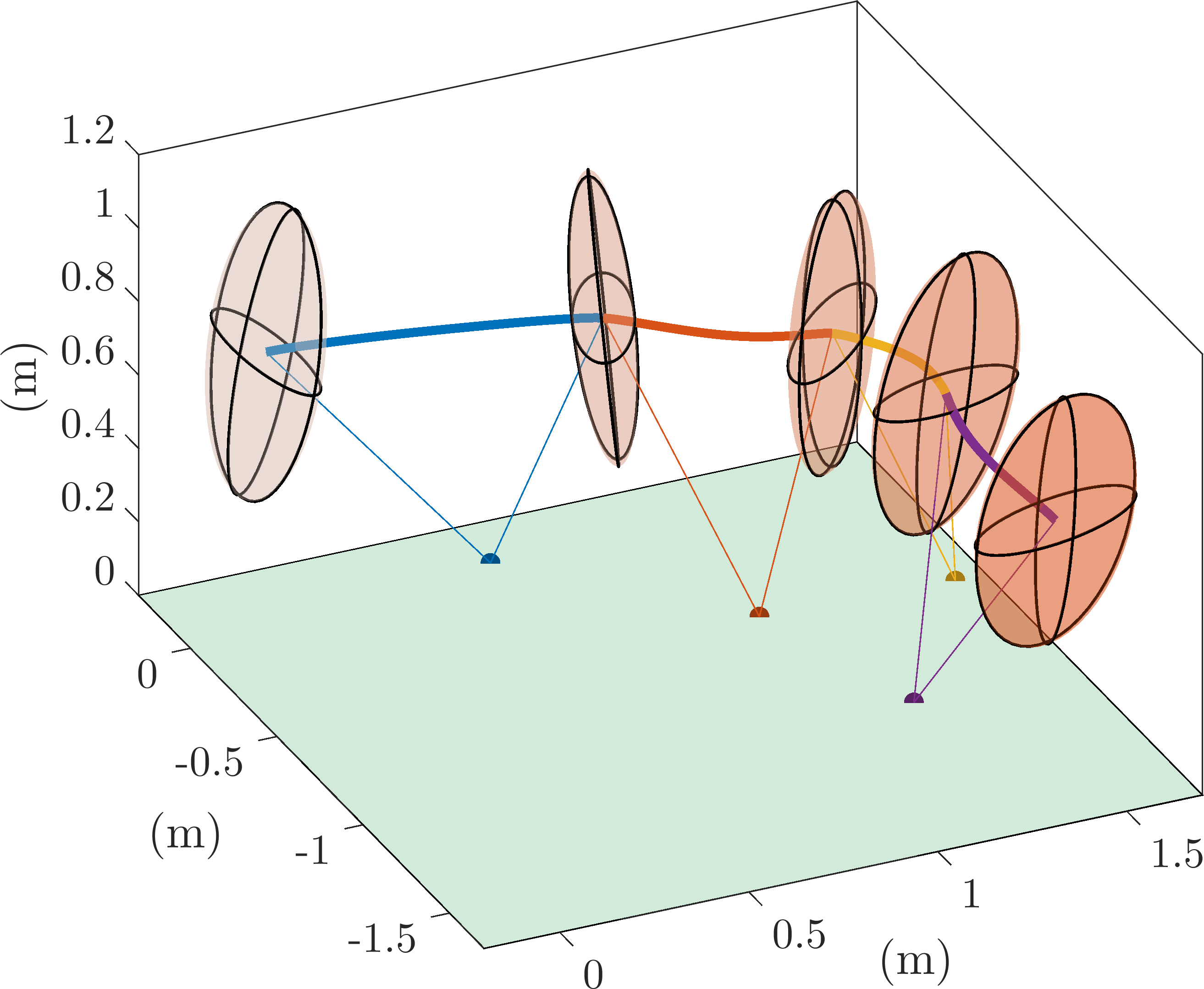}
  \caption{Plot of the reference trajectory for a 2.5 m/s, four-step turn from the optimized single rigid-body model moving left to right. The thick line represents the center of mass path, with different colors showing the different stance phases. Thin lines show leg positions at the start and end of stance phases.}
  \label{fig:turning_reference}
\end{figure}

The resulting library of turn references spans from 0.0 to 2.5 m/s.
The 2.5 m/s turning trajectory is shown in \cref{fig:turning_reference}.
The trajectories have smoothly varying body motions, footstep locations, ground reaction forces, and step timing.

\subsection{Policy Network Design}
The policy architecture used in this work is derived from previous work on applying LSTM networks to bipedal locomotion control \cite{LSTMCassie}. Both actor and critic networks are LSTM RNNs of size 128x128. The state space inputs to our control policy concatenates information from the robot state estimator along with a maneuver progression counter, two periodic clock waveforms, and a target forward heading speed for a total input space size of 42. The breakdown of the state space is shown in \cref{tab:input-table}. From our testing, using a reduced state input set by omitting the maneuver progression counter and pelvis translational acceleration estimates also produces successful turning policies with no noticeable difference in simulation behavior and training time. 

\begin{table}
\centering
\vspace{4pt}
\resizebox{5.8cm}{!}{%
\begin{tabular}{|l|r|}
\hline
\textbf{Policy Input}             & \textbf{Size} \\ \hline
Pelvis Orientation Quaternion                &  4             \\ 
Pelvis Angular Velocity           & 3               \\ 
Pelvis Translational Acceleration & 3              \\ 
Joint Positions and Velocities & 28 \\ 
Maneuver Progression              & 1               \\ 
Clock Signal                      & 2              \\ 
Target Forward Speed              & 1             \\ \hline
\end{tabular}%
}
\caption{The inputs into the learned control policies. All state information is estimated from real or simulated sensor data.}
\label{tab:input-table}
\end{table}

The action space of the policy consists of position targets for all 10 actuated joints on Cassie. The actions are updated at our nominal control rate of 40hz, which are then sent to joint-level PD controllers running at 2 kHz.

\subsection{Reward Function Formulation}\label{reward-method}

%
%
To support an ablation study, we trained policies using different reward functions, \emph{Full Reference, Subset Reference, Foot Timing,} and \emph{No Reference}, each with a different set of additive reward components that capture different aspects of reference information. 
Table \ref{tab:reward_terms} gives the individual component weights for each of the four reward functions. 
All weights are rounded the nearest percentage point. 

Two of the reward components are common across all four reward function variations: 
\begin{itemize}
\item A contact mode reward $r_\text{{contact}}$, which specifies when each foot should be in swing or in stance with a piecewise linear clock function. The gait parameters that define such a function (stepping frequency and swing ratio) and calculated from the reference information.
We refer readers to previous work \cite{jeremyLoad} for further details. 
\item Action smoothness, torque cost, motor velocity costs $r_\text{{ctrl}}$ on the hip roll and yaw motors, and self collision avoidance rewards to promote successful sim-to-real transfer. 
\end{itemize}

The four reward functions are summarized below.  
\subsubsection{Full Reference}
The tracking components of the reward function include pelvis yaw angle ($\psi$), pelvis linear velocity ($\textbf{v}$), pelvis angular momentum ($\textbf{L}$), and the relative distance vector between the pelvis COM and the stance foot ($\textbf{pose}$).
They are given by

\begin{gather}
   r_{\psi}^\text{ref} = \exp{(-\left|(3(\psi_{\text{pelv}} - \psi^{\text{ref}}_{\text{pelv}})\right|)}  \label{eqn:yawR} \\
   r_{\textbf{\text{v}}} = \exp{(-\left\| 2(\textbf{v}_{\text{pelv}} - \textbf{v}^{\text{ref}}_{\text{pelv}}) \right\|_{1} )} \label{eqn:vR} \\
   r_{\textbf{L}} = \exp{(- \left\| \mathbf{L}_{\text{body}} - \mathbf{L}^{\text{ref}}_{\text{body}}\right\| _{1})}  \label{eqn:LR} \\
   r_{\textbf{pose}} = \exp{(- \left\| 5(\textbf{p}_{\text{pose}} - \textbf{p}^{\text{ref}}_{\text{pose}}) \right\|_1 )} \label{eqn:fpR}
\end{gather}



\subsubsection{Subset Reference}
Includes only a subset of full reference rewards, specifically $r_{\psi}^\text{ref}, r_{\mathbf{v_{xy}}}, r_{\mathbf{pose_{xy}}}, r_\text{contact}, r_\text{ctrl}$.
Notably, this omits the angular momentum tracking term in equation \cref{eqn:LR}, as well as tracking only the planar $x,y$ components of $r_\textbf{v}$ and $r_\textbf{pose}$. This particular reward function was chosen because angular momentum tracking was found to have no noticeable effects on the behavior of the resulting policies. 

\subsubsection{Foot Timing}
Omits all tracking rewards \cref{eqn:yawR,eqn:vR,eqn:LR,eqn:fpR} and only consists of $r_\psi^{\text{interp}}, r_{\text{contact}}, r_{\text{ctrl}}$. The only reference information present in the reward is the gait parameters for the contact mode reward term.
$r_\psi^\text{interp}$ replaces $r_\psi^\text{ref}$ and tracks a target that linearly interpolates between 0 and $-\pi/2$ within the timespan of the reference turning maneuvers instead of the optimized yaw trajectory $\psi_\text{pelv}^\text{ref}$.
\subsubsection{No Reference}
A reference-free policy similar to \emph{Foot Timing} that also omits the tracking rewards \cref{eqn:yawR,eqn:vR,eqn:LR,eqn:fpR}. It uses $r_\psi^{\text{interp}}, r_{\text{contact}}, r_{\text{ctrl}}$ exclusively, but in contrast to \emph{Foot Timing}, the gait parameters for $r_{\text{contact}}$ are set by a hand-tuned heuristic. Thus, this policy uses \textbf{no} information from the reference trajectory.

\begingroup
\setlength{\tabcolsep}{8pt} 
\renewcommand{\arraystretch}{1.3} 
\begin{table}
\centering
\vspace{4pt}
\begin{tabular}{|l|r|r|r|r|}
\hline
\textbf{Reward }             & \textbf{Full Ref.}  & \textbf{Sub Ref.}  & \textbf{Foot Timing}  & \textbf{No Ref.} \\ \hline
$r_{\psi}^{\text{ref}}$                &  6     &  3 &  - &  -         \\ 
$r_\psi^{\text{interp}}$                &  -     &  - &  20 &  20         \\ 
$r_{\mathbf{v_{xy}}}$           & 6     &  22 &  - &  -           \\ 
$r_{\mathbf{v_{z}}}$           & 3     &  - &  - &  -           \\
$r_{\mathbf{pose_{xy}}}$                         & 13 &  22 &  -   &  -            \\ 
$r_{\mathbf{pose_{z}}}$                         & 6 &  - &  -   &  -            \\ 
$r_{\mathbf{L}}$ & 9        &  - &  - &  -       \\
$r_{\text{contact}}$ & 38        &  32 &  53 &  53       \\ 
$r_{\text{ctrl}}$ & 19        &  22 &  27 &  27       \\
\hline
\end{tabular}%

\caption{Reward component composition and weighting percentages.}
\label{tab:reward_terms}
\end{table}
\endgroup

\subsection{Episode Initialization}

The beginning of a training episode for turning needs to be reset to a configuration that is a close match to the starting ROM configuration specified by the turning trajectory. 
%
For this purpose, a set of initialization poses $P_{\text{init}}(v,\theta)$ is generated by executing a pre-trained running policy in simulation for a sweep of commanded speeds that match the speeds $v$ of the trajectories in the reference library.
The configurations $[q,\dot{q}]$ of Cassie within a range of gait phases $\theta$ before and after a left-foot swing apex (the starting point of the reference trajectories) are saved to $P_{\text{init}}$. 
On every reset, the configuration state of Cassie is uniformly sampled from the set of poses in $P_{\text{init}}$ for a desired initial speed. 

\subsection{Epilogue Reward}\label{epilogue-method}

Since we want to transition back to walking after executing a turn, it is paramount that the turning policy fulfill the terminal objective of ending in a state that can successfully initialize walking in order to return to a nominal locomotion gait. 
We introduce the novel concept of training with an epilogue reward as a component of our training framework in order to allow turning policies $\pi^{turn}$ to successfully switch back to the nominal locomotion policies $\pi^{walk}$ once the turning policies have reached the end of the reference trajectory states. 

\begin{figure}
  \centering
  \includegraphics[width=0.8\columnwidth]{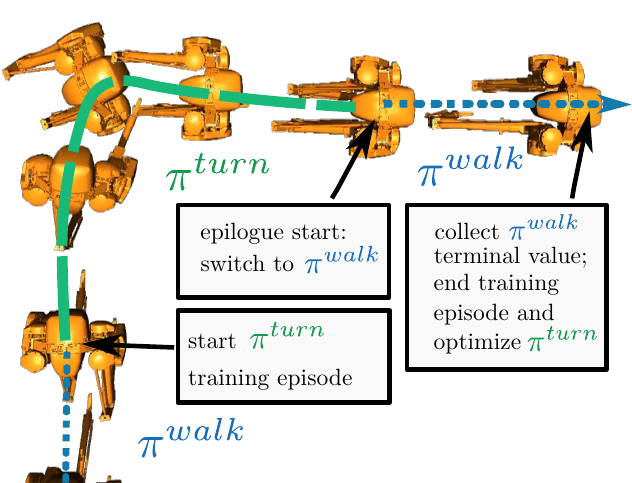}
  \caption{Visualization of a PPO rollout during training. After being initialized from a $\pi^{walk}$ pose, $\pi^{turn}$ is evaluated until the end of the turning maneuver. If $\pi^{turn}$ completed the turning maneuver, $\pi^{walk}$ subsequently takes over to generate the epilogue reward.}
  \label{fig:lead_figure}
\end{figure}

At the end of a PPO rollout, the critic value for the final state $V_{\pi}(S_{T})$ is used as the terminal value in the discounted chain of rewards received at each episode step to estimate the sum of any future discounted rewards \cite{ppo}. 
This is analogous to calculating the $n$-step temporal-difference (TD) returns, where the final value is an estimate for the uncollected rewards beyond the $n$-step horizon \cite{sutton_barto_2020}. 

The epilogue reward introduced in this work modifies the terminal value used at the end of a $\pi^{turn}$ training episode.
It is computed as the discounted sum of returns of the epilogue episode, which triggers when the turning policy has successfully reached the end of its maneuver. In the epilogue episode, the walking policy $\pi^{walk}$ takes over from the last state of the turning episode (normal PPO rollout of the $\pi^{\text{turn}}$ turning policy), and is evaluated deterministically for $k$ simulation steps. 
Once the epilogue is complete, the epilogue reward becomes the value of $V_{\pi^{\text{turn}}}(S_{T})$ used to optimize $\pi^{\text{turn}}$. 
Formally, the epilogue reward is
\begin{equation}
V_{\pi^{\text{turn}}}(S_{T}) = \Big( \sum\limits_{i=T}^{T+k}\gamma^{i-T}R_{i+1} \Big) + \gamma^{T+k}V_{\pi^{\text{walk}}}(S_{T+k+1}) 
\end{equation}
where $k$ is the length of the epilogue, $T$ is the length of the turning maneuver, $R_{i+1}$ is based on the reward function used to traing $\pi^{\text{walk}}$, and $V_{\pi^{\text{walk}}}$ is the critic trained for the walking policy. 
Modifying the estimate of future returns in this manner incentivizes $\pi^{turn}$ to terminate in a configuration $[q, \dot{q}]$ amenable for the execution of $\pi^{walk}$ by maximizing the epilogue returns for continued walking.  
As a control for the epilogue, a \emph{No Epilogue} policy is trained using the exact same rewards as \emph{Foot Timing}, but with $k$ set to 0. The \emph{Foot Timing} reward set was chosen over the others because it produced the highest quality turning behaviors in sim as well as fast convergence times. 
A value of 120 is used for $k$ for all other policies.

\subsection{Dynamics Randomization}
We applied dynamics randomization as described in \cite{LSTMCassie} during the training process of our turning controller to help close the sim-to-real gap and enable a successful transfer to real hardware. In addition, we also apply a constant perturbance force to the robot pelvis over the course of a training episode with a randomly sampled magnitude and direction in order to promote the emergence of robust turning behavior. 
The details of our randomization parameters can be found in \cref{tab:dynamics-rand}. 

\begin{table}
\vspace{6pt}
\resizebox{\columnwidth}{!}{%
\begin{tabular}{|l|l l|}
\hline
\textbf{Parameter}           & \textbf{Range}            & \textbf{Unit} \\ \hline
Policy Control Rate              & {[}0.95,1.05{] $\times$ default}           & Hz             \\ 
Joint Encoder Noise          & {[}-0.05, 0.05{]}         & rad           \\ 
Joint Damping                & {[}0.8, 2.5{] $\times$ default}            & Nms/rad       \\ 
Link Mass                    & {[}0.9, 1.5{] $\times$ default}            & kg            \\ 
Friction Coefficient         & {[}0.45, 1.3{]}           & -             \\ 
External Force Magnitude & {[}0, 40{]}               & N             \\ 
External Force Dir. (Azimuth)   & {[}0, 2$\pi${]}           & rad           \\ 
External Force Dir. (Elevation) & {[}0, $\frac{\pi}{4}${]} & rad           \\ 
Initial Pelvis Velocity (x)      & {[}-0.3,0.3{]} + default  & m/s           \\ Initial Pelvis Velocity (y)      & {[}-0.4,0.4{]}            & m/s           \\ \hline
\end{tabular}%
}
\caption{Randomization Range Parameters}
\label{tab:dynamics-rand}
\end{table}
\section{Results}

To evaluate the utility and necessity of our optimized ROM trajectories and the epilogue reward, we assess and compare the set of policies proposed in \cref{reward-method} and \cref{epilogue-method} in simulation for their performance and turning behavior characteristics. 
We also present successful sim-to-real transfer of a selection of the policies tested in simulation in our submission video.

\subsection{Simulation Results}
\subsubsection{Sample Efficiency} 
\begin{figure}
  \centering
  \includegraphics[width=0.78\columnwidth]{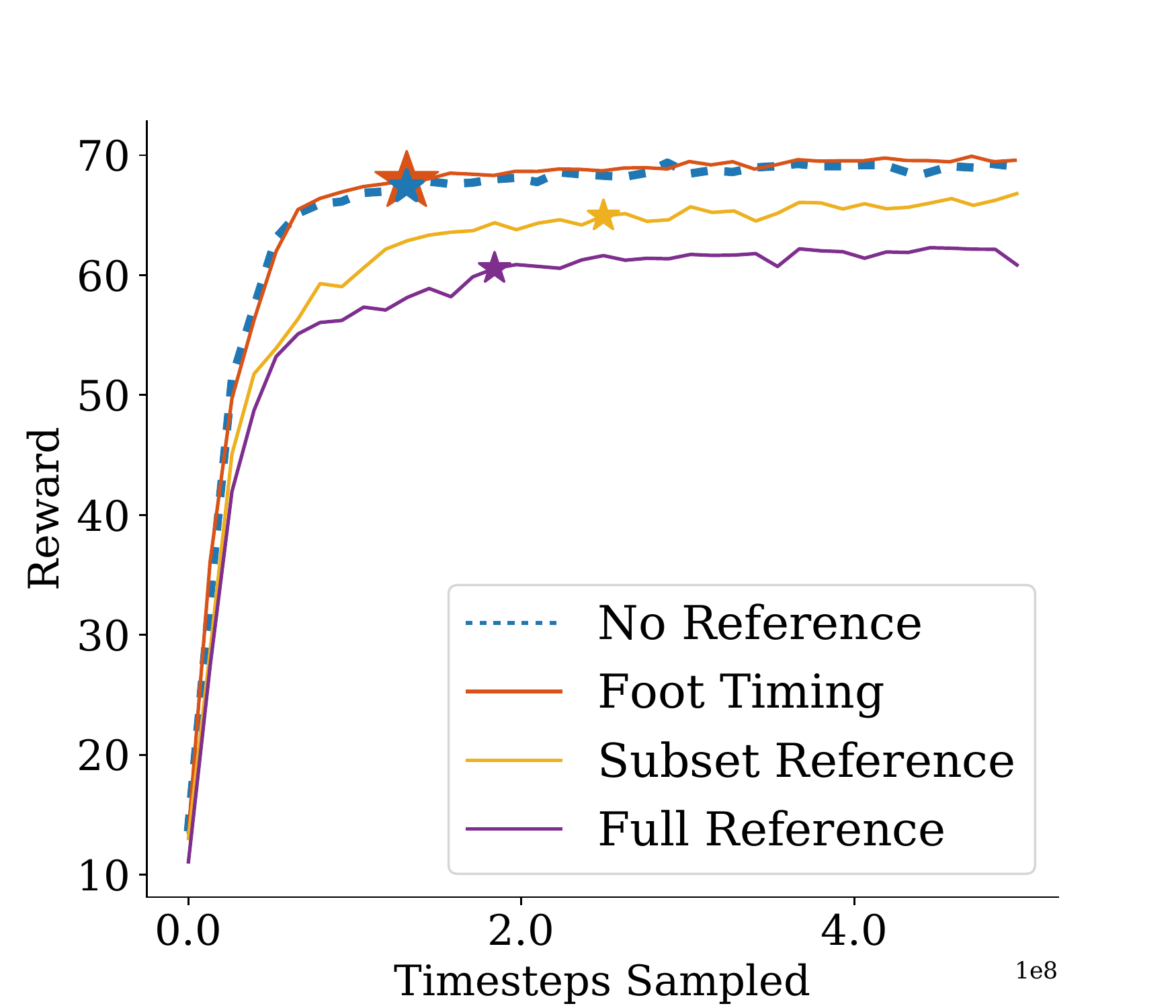}
  \caption{Comparison of sample efficiency for our proposed turning policies. Note that the absolute scale of the different curves are not necessarily comparable since each reward function include different reward components. The star symbols mark the time to convergence for each policy, which is the point on the learning curve that exceeds 97$\%$ of the maximum reward seen during training for the first time.}
  \label{fig:learning}
\end{figure}

We plot the learning curves for each policy in Figure \cref{fig:learning} to compare the sample efficiency of our turning policies. 
Since each policy is trained with different reward functions, the reward values attained by each policy can not be used to form conclusions about their relative performance. 
Instead, we compare policies by the number of samples to convergence, shown for each turning policy by star symbols in Figure \cref{fig:learning} that mark when each policy first surpassed 97$\%$ of the maximum reward value experienced during training. Notably, the policies that use less information from the optimized trajectories converge slightly faster than the policies that follow the SRBM reference trajectory more faithfully. We hypothesize that the learning speed disparity may be attributed to model differences between the SRBM and Cassie's full-order dynamics leading to conflicting interactions between the tracking reward terms. This may cause policies that track more of the reference data to require more samples in order to learn to optimize for multiple conflicting objectives before convergence. 

\subsubsection{Turning Behavior}

\begin{figure}
  \centering
  \vspace{4 pt}
  \includegraphics[width=0.90\columnwidth]{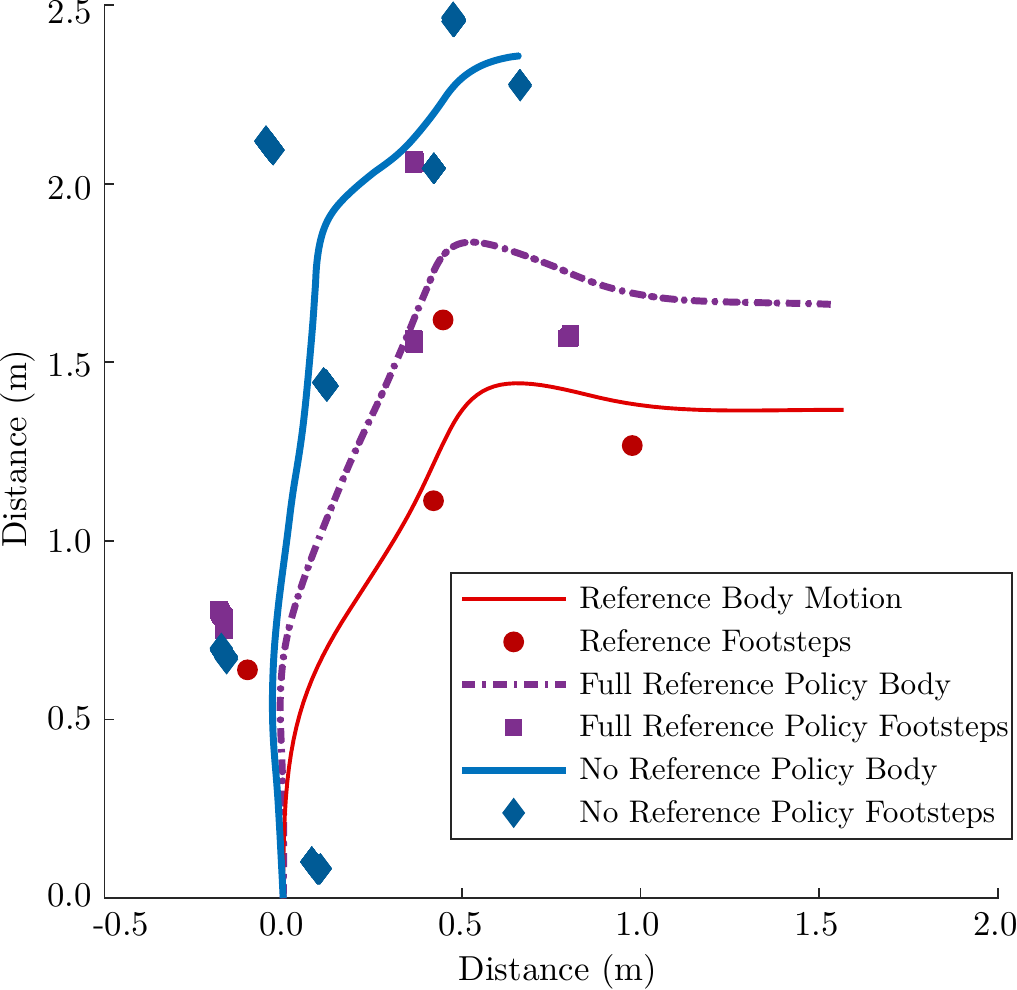}
  \caption{Plot of footstep touchdown locations and pelvis trajectory for the reference data, \emph{Full Reference} and \emph{No Reference} policies for a turning maneuver executed at 2.5m/s. }
  \label{fig:feet_fullRef}
\end{figure}

\begin{figure}
  \centering
  \vspace{4 pt}
  \includegraphics[width=0.82\columnwidth]{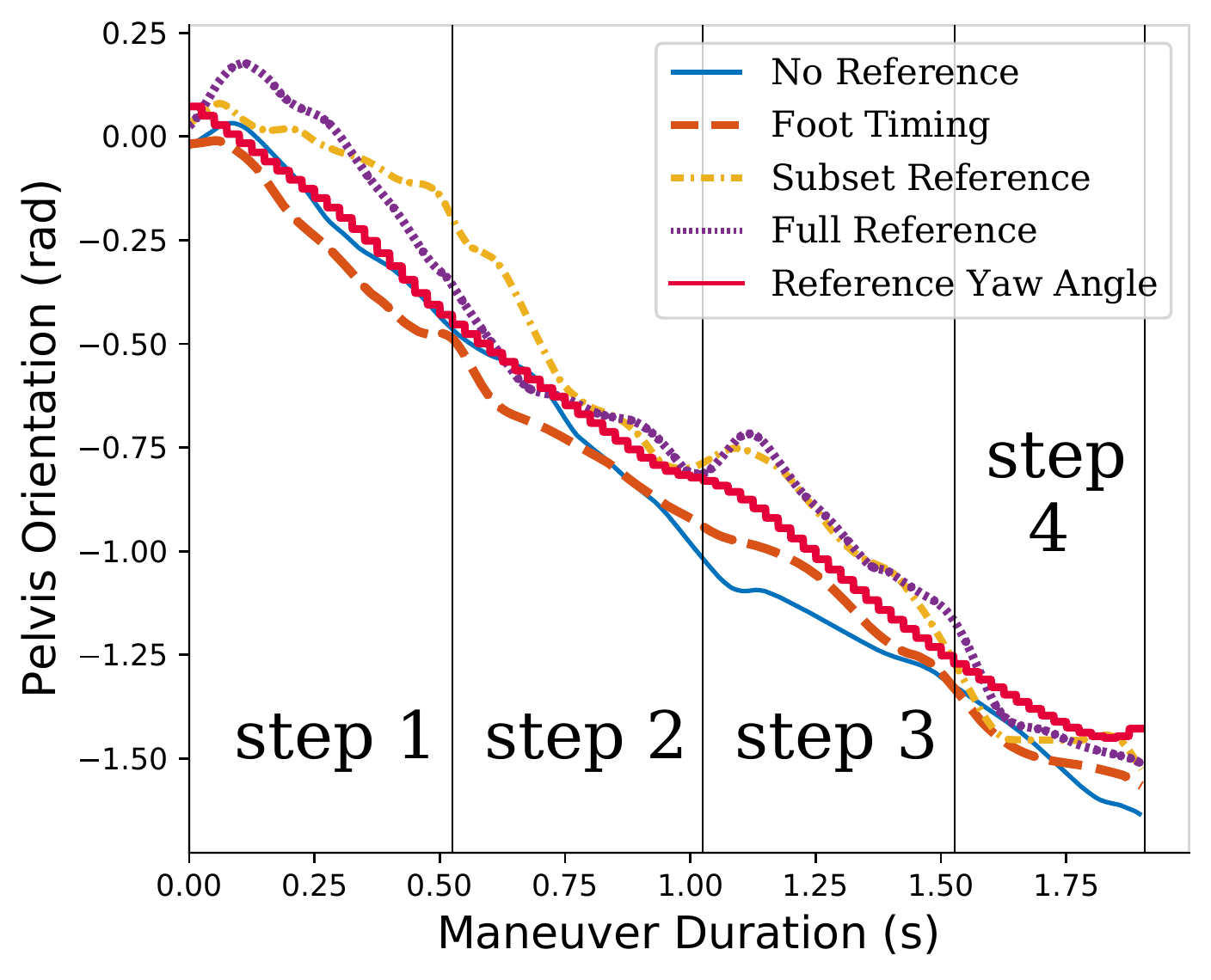}
  \caption{Simulated pelvis yaw angles for various turning policies over the course of a turning maneuver. The Foot Timing and No Reference policies are not trained trained to track the reference data yaw trajectory shown in solid red.}
  \label{fig:yawTrajectory}
\end{figure}

\cref{fig:feet_fullRef} compares the turning trajectory of the \emph{Full Reference} and \emph{No Reference} policies for a single sample trial in simulation against the trajectory prescribed by the reference data. 
Since the \emph{No Reference} policy is trained to match a footstep contact schedule set by a heuristic instead of following the trajectory data, it completes the 90\degree{} turn in seven steps rather than four. 
As a result, the pelvis trajectory and footstep placements distinctly differs from the reference data since it is trained to not track the reference data. 
This is in contrast to the \emph{Full Reference} policy turning behavior, where the features of the pelvis trajectory is similar to that of the reference data, and the placement of its stance feet relative to the body also closely match those of the reference. 
Policies trained on subsets of the tracking rewards all produce four-step turning behaviors similar to the results of the \emph{Full Reference} policy, indicating that the only necessary reference trajectory information for training policies to perform four-step 90\degree{} turns is a feasible footstep contact schedule. 
\cref{fig:yawTrajectory} illustrates the change in orientation of the robot pelvis over the course of a turning maneuver, which is not communicated by the pelvis COM trajectories illustrated in \cref{fig:feet_fullRef}. The \emph{Full Reference} and \emph{Subset Reference} policies are the two policies rewarded to track the optimized body yaw angle trajectory, but we see observe noticeable deviations from the target yaw trajectories at the beginning of the first and third footsteps. This is likely caused by the policies learning to maximize rewards of multiple conflicting objectives from the reference data, such as pelvis linear velocity and body yaw angles.

\subsubsection{Policy Robustness}

\begin{figure}
  \centering
  \vspace{6 pt}
  \includegraphics[width=0.98\columnwidth]{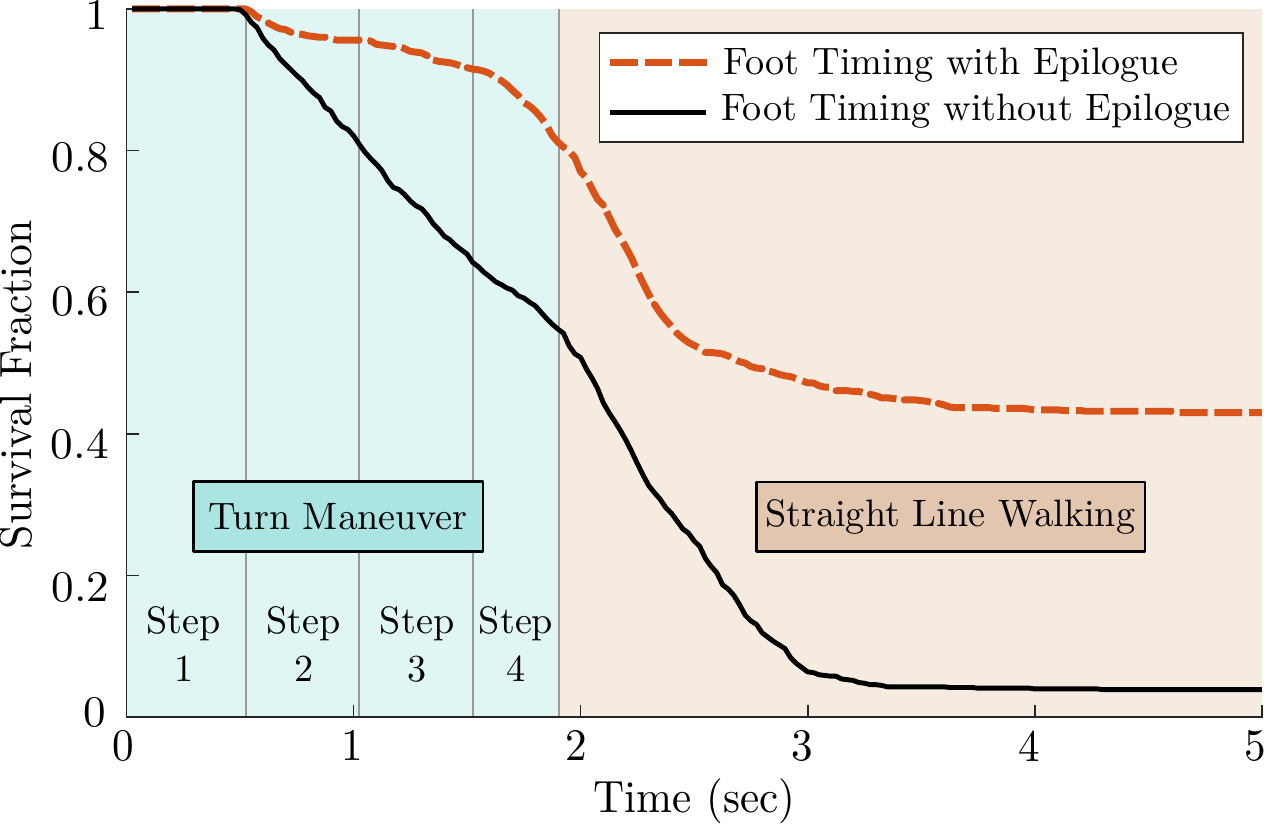}
  \caption{Robustness comparison between the \emph{Foot Timing} policy trained with epilogue and the \emph{Foot Timing} policy trained without using the same method as  \cref{fig:robustness_reference}.}
  \label{fig:robustness_epilogue}
\end{figure}

\begin{figure}
  \centering
  \vspace{4 pt}
  \includegraphics[width=0.98\columnwidth]{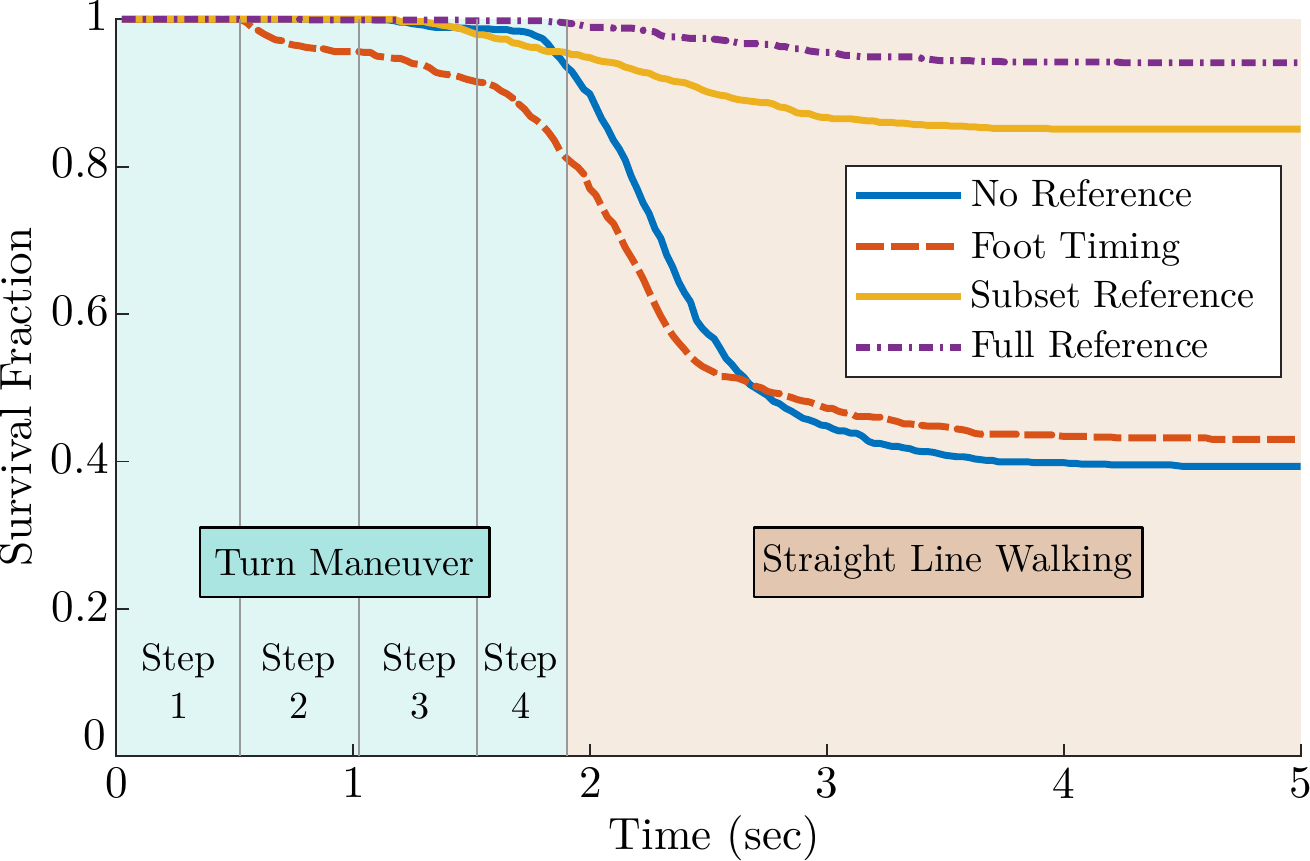}
  \caption{Robustness comparison between our proposed turning policies conducted for 1000 turning maneuver trials at 2.5m/s. The step labels denote the reference data step progression timings produced by TO. Since the \emph{No Reference} policy is trained to follow a contact schedule set by a heuristic, the step labels do not apply to this policy.}
  \label{fig:robustness_reference}
\end{figure}

We simulate 1000 trials of 2.5m/s turning maneuvers for each turning policy to assess its ability to complete a turn and switch back to $\pi^{walk}$ successfully in the presence of a constant perturbance force applied to the body during the execution of the turning maneuver. 
A random direction is sampled before each turning maneuver trial, and a constant force of 35N is applied in the chosen direction. 
The time at which the the policy falls over is logged for each trial to produce the policy survival plots shown in \cref{fig:robustness_reference,fig:robustness_epilogue}. 
From \cref{fig:robustness_reference}, policies trained with more of the reference trajectory data is more robust at rejecting perturbance forces, with the exception that the \emph{No Reference} policy outperforms the \emph{Foot Timing} policy during the turn maneuver. 
Since all policies experience perturbance forces during training, it is possible that rewarding policies to track certain elements of the trajectory data such as foot placement positions and pelvis translational velocities are beneficial to policy robustness. 
\cref{fig:robustness_epilogue} compares the effects of training with and without the use of epilogue rewards. 
While the \emph{Foot Timing} policy achieved a survival fraction of around 50$\%$, the \emph{No Epilogue} policy fared significantly worse than its counterpart with a terminal survival fraction of just 4$\%$. We observed that \emph{No Epilogue} policy had a low transition success rate, which indicates that epilogue rewards are necessary for reliable switching between $\pi^{\text{turn}}$ and $\pi^{\text{walk}}$ policies. We hypothesize that using other reward functions without the epilogue reward will produce similar results, but did not run such tests in this work.

\subsection{Hardware Results}

During our outdoors hardware tests, we were able to demonstrate successful turning maneuvers and policy switching on artificial turf with the \emph{No Reference} and \emph{Full Reference} policies. The \emph{Subset Reference} policy was also tested, but we were unable to switch back to the walking policy without falling.
We also successfully tested the \emph{Foot Timing} policy indoors on multiple low speed turns performed in succession. 
While our simulation results indicate that the \emph{Full Reference} policy should perform more consistently than the \emph{No Reference} policy on real hardware, we observed that the \emph{No Reference} policy was more consistent than the \emph{Full Reference} policy at turning and transitioning in our outdoors tests. 
The \emph{Full Reference} and \emph{Subset Reference} policies pitch the pelvis during the turn more so than the \emph{No Reference} and \emph{Foot Timing} policies that in contrast keep the pelvis fairly level throughout the turn. 
This is consistent with what we see in simulation, although the \emph{Full Reference} pelvis pitching motion in simulation is more fluid and intricate than our corresponding hardware results which maintain an awkward downward pitch throughout the entire turn. 
Due to unresolved sim-to-real challenges, we were unable to replicate the same consistent performance seen in simulation in our hardware trials. 
From our limited hardware tests, the \emph{No Reference} policy seemed to have a higher chance at executing successful turns than the \emph{Full Reference} policy. We refer readers to the attached video for full hardware results.

\section{CONCLUSIONS}

In this work, we present a novel learning framework to generate aperiodic behaviors using SRBM trajectories on the bipedal robot Cassie. 
Epilogue rewards are a key component of our learning framework to facilitate smooth transitions between nominal locomotion policies and the policy trained to execute the desired maneuver.   
While our methods exhibited promising results in simulation, we encountered difficulties with sim-to-real and were unable to fully transfer the success of our simulation results to hardware field trials. 
Although we were able to demonstrate repeated successful turning maneuvers at low speeds on hardware, turning maneuvers committed at faster walking speeds were much less reliable. 
A possible reason for the performance gap might stem from issues with the hardware state estimator producing inaccurate orientation estimates only when large pelvis accelerations are experienced during the execution of turning maneuvers at higher speeds. Our reference-based policies command much larger pelvis pitch angles over the course of a turn than the $\emph{No Reference}$ and $\emph{Foot Timing}$ policies which may have exacerbated the state estimation problems. 
It is possible that this difference may be a contributing factor to why the $\emph{No Reference}$ policy performed better than the reference-based policies on hardware when our simulation results suggest the opposite.
One downside of the learning framework introduced in this work is the challenge of scaling to switching between multiple aperiodic behaviors sourced from a diverse behavior trajectory library. 
Future avenues of research could build upon this work by investigating how to effectively switch between large sets of individual behavior policies in order to allow for the execution of more complex dynamic routines such as dancing or parkour. 

\addtolength{\textheight}{-0cm}   







\def\bibfont{\footnotesize}
\bibliographystyle{IEEEtran}
\bibliography{main}

\begin{thebibliography}{10}
\providecommand{\url}[1]{#1}
\csname url@rmstyle\endcsname
\providecommand{\newblock}{\relax}
\providecommand{\bibinfo}[2]{#2}
\providecommand\BIBentrySTDinterwordspacing{\spaceskip=0pt\relax}
\providecommand\BIBentryALTinterwordstretchfactor{4}
\providecommand\BIBentryALTinterwordspacing{\spaceskip=\fontdimen2\font plus
\BIBentryALTinterwordstretchfactor\fontdimen3\font minus
  \fontdimen4\font\relax}
\providecommand\BIBforeignlanguage[2]{{%
\expandafter\ifx\csname l@#1\endcsname\relax
\typeout{** WARNING: IEEEtran.bst: No hyphenation pattern has been}%
\typeout{** loaded for the language `#1'. Using the pattern for}%
\typeout{** the default language instead.}%
\else
\language=\csname l@#1\endcsname
\fi
#2}}

\bibitem{DesignDynLegRob}
\BIBentryALTinterwordspacing
S.~Kim and P.~M. Wensing, ``Design of dynamic legged robots,''
  \emph{Foundations and Trends® in Robotics}, vol.~5, no.~2, pp. 117--190,
  2017. [Online]. Available: \url{http://dx.doi.org/10.1561/2300000044}
\BIBentrySTDinterwordspacing

\bibitem{cheetahAutoJump}
\BIBentryALTinterwordspacing
H.-W. Park, P.~M. Wensing, and S.~Kim, ``Jumping over obstacles with mit
  cheetah 2,'' \emph{Robotics and Autonomous Systems}, vol. 136, p. 103703,
  2021. [Online]. Available:
  \url{https://www.sciencedirect.com/science/article/pii/S0921889020305431}
\BIBentrySTDinterwordspacing

\bibitem{backflip_mini}
B.~Katz, J.~D. Carlo, and S.~Kim, ``Mini cheetah: A platform for pushing the
  limits of dynamic quadruped control,'' in \emph{2019 International Conference
  on Robotics and Automation (ICRA)}, 2019, pp. 6295--6301.

\bibitem{all_gaits}
J.~Siekmann, Y.~Godse, A.~Fern, and J.~Hurst, ``Sim-to-real learning of all
  common bipedal gaits via periodic reward composition,'' in \emph{2021 IEEE
  International Conference on Robotics and Automation (ICRA)}, 2021, pp.
  7309--7315.

\bibitem{blind_stairs}
J.~Siekmann, K.~Green, J.~Warila, A.~Fern, and J.~Hurst, ``{Blind Bipedal Stair
  Traversal via Sim-to-Real Reinforcement Learning},'' in \emph{Proceedings of
  Robotics: Science and Systems}, Virtual, July 2021.

\bibitem{Batke2022SRBM}
\BIBentryALTinterwordspacing
R.~Batke, F.~Yu, J.~Dao, J.~Hurst, R.~L. Hatton, A.~Fern, and K.~Green,
  ``Optimizing bipedal maneuvers of single rigid-body models for reinforcement
  learning,'' 2022. [Online]. Available: \url{https://arxiv.org/abs/2207.04163}
\BIBentrySTDinterwordspacing

\bibitem{AgileSim2Real}
\BIBentryALTinterwordspacing
J.~Tan, T.~Zhang, E.~Coumans, A.~Iscen, Y.~Bai, D.~Hafner, S.~Bohez, and
  V.~Vanhoucke, ``{Sim-to-Real: Learning Agile Locomotion For Quadruped
  Robots},'' in \emph{Robotics: Science and Systems XIV}.\hskip 1em plus 0.5em
  minus 0.4em\relax Pittsburgh, Pennsylvania: Robotics: Science and Systems
  Foundation, jun 2018. [Online]. Available:
  \url{http://www.roboticsproceedings.org/rss14/p10.html}
\BIBentrySTDinterwordspacing

\bibitem{anymalRL}
\BIBentryALTinterwordspacing
J.~Lee, J.~Hwangbo, L.~Wellhausen, V.~Koltun, and M.~Hutter, ``Learning
  quadrupedal locomotion over challenging terrain,'' \emph{Science Robotics},
  vol.~5, no.~47, p. eabc5986, 2020. [Online]. Available:
  \url{https://www.science.org/doi/abs/10.1126/scirobotics.abc5986}
\BIBentrySTDinterwordspacing

\bibitem{helei_stepping_stones}
\BIBentryALTinterwordspacing
H.~Duan, A.~Malik, M.~S. Gadde, J.~Dao, A.~Fern, and J.~Hurst, ``Learning
  dynamic bipedal walking across stepping stones,'' in \emph{accepted to 2022
  IEEE/RSJ International Conference on Intelligent Robotics and Systems}, 2022.
  [Online]. Available: \url{https://arxiv.org/abs/2205.01807}
\BIBentrySTDinterwordspacing

\bibitem{LSTMCassie}
J.~Siekmann, S.~Valluri, J.~Dao, F.~Bermillo, H.~Duan, A.~Fern, and J.~Hurst,
  ``{Learning Memory-Based Control for Human-Scale Bipedal Locomotion},'' in
  \emph{Proceedings of Robotics: Science and Systems}, Corvalis, Oregon, USA,
  July 2020.

\bibitem{DeepMindLearnCoreSkills}
\BIBentryALTinterwordspacing
R.~Hafner, T.~Hertweck, P.~Kl{\"{o}}ppner, M.~Bloesch, M.~Neunert,
  M.~Wulfmeier, S.~Tunyasuvunakool, N.~Heess, and M.~A. Riedmiller, ``Towards
  general and autonomous learning of core skills: {A} case study in
  locomotion,'' in \emph{4th Conference on Robot Learning, CoRL 2020, 16-18
  November 2020, Virtual Event / Cambridge, MA, {USA}}, vol. 155.\hskip 1em
  plus 0.5em minus 0.4em\relax {PMLR}, 2020, pp. 1084--1099. [Online].
  Available: \url{https://proceedings.mlr.press/v155/hafner21a.html}
\BIBentrySTDinterwordspacing

\bibitem{deepmind_learn_locomotion_rich_envs}
\BIBentryALTinterwordspacing
N.~Heess, D.~TB, S.~Sriram, J.~Lemmon, J.~Merel, G.~Wayne, Y.~Tassa, T.~Erez,
  Z.~Wang, S.~M.~A. Eslami, M.~Riedmiller, and D.~Silver, ``Emergence of
  locomotion behaviours in rich environments,'' 2017. [Online]. Available:
  \url{https://arxiv.org/abs/1707.02286}
\BIBentrySTDinterwordspacing

\bibitem{cfrost}
A.~Hereid, O.~Harib, R.~Hartley, Y.~Gong, and J.~W. Grizzle, ``Rapid trajectory
  optimization using c-frost with illustration on a cassie-series dynamic
  walking biped,'' in \emph{2019 IEEE/RSJ International Conference on
  Intelligent Robots and Systems (IROS)}, 2019, pp. 4722--4729.

\bibitem{posa}
\BIBentryALTinterwordspacing
M.~Posa, C.~Cantu, and R.~Tedrake, ``A direct method for trajectory
  optimization of rigid bodies through contact,'' \emph{The International
  Journal of Robotics Research}, vol.~33, no.~1, pp. 69--81, 2014. [Online].
  Available: \url{https://doi.org/10.1177/0278364913506757}
\BIBentrySTDinterwordspacing

\bibitem{Kuindersma2016OptimizationbasedLP}
S.~Kuindersma, R.~Deits, M.~F. Fallon, A.~K. Valenzuela, H.~Dai, F.~Permenter,
  T.~Koolen, P.~Marion, and R.~Tedrake, ``Optimization-based locomotion
  planning, estimation, and control design for the atlas humanoid robot,''
  \emph{Autonomous Robots}, vol.~40, pp. 429--455, 2016.

\bibitem{DaiWBC_centroidal}
H.~Dai, A.~Valenzuela, and R.~Tedrake, ``Whole-body motion planning with
  centroidal dynamics and full kinematics,'' in \emph{2014 IEEE-RAS
  International Conference on Humanoid Robots}, 2014, pp. 295--302.

\bibitem{Apgar2018FastOT}
T.~Apgar, P.~Clary, K.~Green, A.~Fern, and J.~W. Hurst, ``Fast online
  trajectory optimization for the bipedal robot cassie,'' \emph{Robotics:
  Science and Systems XIV}, 2018.

\bibitem{QuanJump}
Q.~Nguyen, M.~J. Powell, B.~Katz, J.~D. Carlo, and S.~Kim, ``Optimized jumping
  on the mit cheetah 3 robot,'' in \emph{2019 International Conference on
  Robotics and Automation (ICRA)}, 2019, pp. 7448--7454.

\bibitem{hutterComp}
\BIBentryALTinterwordspacing
M.~Bjelonic, R.~Grandia, M.~Geilinger, O.~Harley, V.~S. Medeiros, V.~Pajovic,
  E.~Jelavic, S.~Coros, and M.~Hutter, ``Offline motion libraries and online
  mpc for advanced mobility skills,'' \emph{The International Journal of
  Robotics Research}, June 2022. [Online]. Available:
  \url{https://doi.org/10.1177/02783649221102473}
\BIBentrySTDinterwordspacing

\bibitem{BD_parkour}
C.~Hennick, ``Leaps, bounds, and backflips,''
  \url{https://blog.bostondynamics.com/atlas-leaps-bounds-and-backflips}, 2021,
  accessed: 2022-05-19.

\bibitem{DeepGait}
V.~Tsounis, M.~Alge, J.~Lee, F.~Farshidian, and M.~Hutter, ``Deepgait: Planning
  and control of quadrupedal gaits using deep reinforcement learning,''
  \emph{IEEE Robotics and Automation Letters}, vol.~5, no.~2, pp. 3699--3706,
  2020.

\bibitem{CCROC}
P.~Fernbach, S.~Tonneau, O.~Stasse, J.~Carpentier, and M.~Taïx, ``C-croc:
  Continuous and convex resolution of centroidal dynamic trajectories for
  legged robots in multicontact scenarios,'' \emph{IEEE Transactions on
  Robotics}, vol.~36, no.~3, pp. 676--691, 2020.

\bibitem{deepmind_centroidal}
\BIBentryALTinterwordspacing
P.~Brakel, S.~Bohez, L.~Hasenclever, N.~Heess, and K.~Bousmalis, ``Learning
  coordinated terrain-adaptive locomotion by imitating a centroidal dynamics
  planner,'' 2021. [Online]. Available: \url{https://arxiv.org/abs/2111.00262}
\BIBentrySTDinterwordspacing

\bibitem{ppo}
\BIBentryALTinterwordspacing
J.~Schulman, F.~Wolski, P.~Dhariwal, A.~Radford, and O.~Klimov, ``Proximal
  policy optimization algorithms,'' 2017. [Online]. Available:
  \url{https://arxiv.org/abs/1707.06347}
\BIBentrySTDinterwordspacing

\bibitem{jeremyLoad}
J.~Dao, K.~Green, H.~Duan, A.~Fern, and J.~Hurst, ``Sim-to-real learning for
  bipedal locomotion under unsensed dynamic loads,'' in \emph{2022
  International Conference on Robotics and Automation (ICRA)}, 2022.

\bibitem{coalesce}
M.~S. Jones, ``Optimal control of an underactuated bipedal robot,'' Master's
  thesis, Oregon State University, 2014.

\bibitem{sutton_barto_2020}
R.~S. Sutton and A.~G. Barto, \emph{Reinforcement learning: An
  introduction}.\hskip 1em plus 0.5em minus 0.4em\relax The MIT Press, 2020.

\end{thebibliography}

\end{document}